\documentclass{article}

\usepackage[numbers]{natbib}
\usepackage[final]{neurips_data_2021}

\usepackage[utf8]{inputenc} % allow utf-8 input
\usepackage[T1]{fontenc}    % use 8-bit T1 fonts
\usepackage[hyphens]{url}            % simple URL typesetting
\usepackage{hyperref}       % hyperlinks
\usepackage{booktabs}       % professional-quality tables
\usepackage{amsfonts}       % blackboard math symbols
\usepackage{nicefrac}       % compact symbols for 1/2, etc.
\usepackage{microtype}      % microtypography
\usepackage{xcolor}         % colors
\usepackage{enumitem}
\usepackage{cleveref}
\usepackage{multirow}
\usepackage{graphicx}
\usepackage{tablefootnote}
\usepackage{float}
\usepackage{caption}

\usepackage{svg}

 \newcommand{\new}[1]{\textcolor{black}{{#1}}}

\usepackage[toc,page,header]{appendix}
\usepackage{minitoc}

\usepackage{array}
\newcolumntype{P}[1]{>{\centering\arraybackslash}p{#1}}
\newcolumntype{C}[1]{>{\centering\let\newline\\\arraybackslash\hspace{0pt}}m{#1}}

\newcommand*\rot{\rotatebox{50}}

\newcommand*{\msceleb}{MS-Celeb-1M}
\newcommand*{\duke}{DukeMTMC}
\newcommand*{\lfw}{Labeled Faces in the Wild}

\title{Mitigating Dataset Harms Requires Stewardship: \\ Lessons from 1000 Papers}

\author{%
  Kenny Peng, Arunesh Mathur, Arvind Narayanan
    \\
  Princeton University\\
}

\begin{document}

\doparttoc % Tell to minitoc to generate a toc for the parts
\faketableofcontents % Run a fake tableofcontents command for the partocs

\maketitle

\begin{abstract}
  Machine learning datasets have elicited concerns about privacy, bias, and unethical applications, leading to the retraction of prominent datasets such as DukeMTMC, MS-Celeb-1M, and Tiny Images.
  In response, the machine learning community has called for higher ethical standards in dataset creation.
  To help inform these efforts, we studied three influential but ethically problematic face and person recognition datasets---Labeled Faces in the Wild (LFW), MS-Celeb-1M, and DukeMTMC---by analyzing nearly 1000 papers that cite them.
  We found that the creation of
derivative datasets and models, broader technological and social change, the lack of clarity of licenses, and
dataset management practices can introduce a wide range of ethical concerns.
  We conclude by suggesting a distributed approach to harm mitigation that considers the entire life cycle of a dataset.
\end{abstract}

\section{Introduction}

Datasets play an essential role in machine learning research but also raise ethical concerns. These concerns include the privacy of individuals included~\cite{megapixels, pyrrhic}, representational harms introduced by annotations~\cite{excavatingai, Hanna2020TowardsAC}, effects of biases on downstream use~\cite{Buolamwini2018GenderSI, Caliskan2017SemanticsDA, Blodgett2017RacialDI}, and use for ethically dubious purposes~\cite{megapixels, solon_19AD, murgia_2019}. These concerns have led to the retractions of prominent research datasets including Tiny Images~\cite{tinyimages}, VGGFace2~\cite{vggface}, \duke~\cite{dukemtmc}, and \msceleb~\cite{ms-celeb-1m}.

The machine learning community has responded to these concerns and has developed ways to mitigate harms associated with datasets. Researchers have worked to make sense of ethical considerations involved in dataset creation~\cite{ethicalhighlighter, paullada2020data, contesting-datasets}, have proposed ways to identify and mitigate biases in datasets~\cite{aifairness360, revise}, have developed means to protect the privacy of individuals in datasets~\cite{pyrrhic, Yang2021ASO}, and have improved methods to document datasets~\cite{datasheets, Holland2018TheDN, nlp-data-statements, Mitchell2019ModelCF}.
% While successful, these efforts have only addressed the range of ethical issues that arise when a dataset is first created. As a result, they inadequately address the concerns that arise \emph{after} a dataset is released and used in practice by the research community.

The premise of our work is that these efforts can be more effective if informed by an understanding of how datasets are used in practice. We present an account of the life cycles of three popular face and person recognition datasets: \lfw{} (LFW)~\cite{lfw}, \msceleb{}~\cite{ms-celeb-1m}, and \duke{}~\cite{dukemtmc}. These datasets have been the subject of recent ethical scrutiny~\cite{ethicalhighlighter} and, in the case of \msceleb{} and \duke{}, have been retracted by their creators. Analyzing nearly 1,000 papers that cite these datasets and their derivative datasets or pre-trained models, we present five findings that describe ethical considerations arising beyond dataset creation:

\begin{itemize}[leftmargin=*]
\item Dataset retraction has a limited effect on mitigating harms (Section \ref{issue:decentralized}). Our analysis shows that even after \duke{} and \msceleb{} were retracted, their underlying data remained widely available and continued to be used in research papers. Because of such ``runaway data,'' retractions are unlikely to cut off data access; moreover, without a clear indication of the underlying intention, retractions may have limited normative influence.

\item Derivatives raise new ethical concerns (Section \ref{issue:derivatives}). %Across \duke{}, \msceleb{}, and LFW, we identified 35 derivative datasets and 6 classes of pre-trained models.
The derivatives of \duke{}, \msceleb{}, and LFW that we document enable the use of the dataset in production settings, introduce new annotations of the data, or apply additional data processing steps. Each of these alterations lead to a unique set of ethical considerations. %Thus, the impact of a dataset may be much broader than its original intention.

\item Licenses, a primary mechanism governing dataset use, can lack substantive effect (Section \ref{issue:tos}). We found that the licenses of \duke{}, \msceleb{}, and LFW do not effectively restrict production use of the datasets. In particular, while the original license of \msceleb{} only permits non-commercial research use of the dataset, only 3 of 21 GitHub repositories we found containing models pre-trained on \msceleb{} included the same designation. We found anecdotal evidence suggesting that production use of models trained on non-commercial datasets is commonplace.

\item The ethical concerns associated with a dataset can change over time, as a result of both technological and social change (Section \ref{issue:context}). In the case of LFW and the influential ImageNet dataset \cite{imagenet}, technological advances opened the door for production use of the datasets, raising new ethical concerns. Additionally, various social factors led to a more critical understanding of the demographic composition of LFW and the annotation practices underlying ImageNet.

\item While dataset management and citation practices can support harm mitigation, current practices have several shortcomings (Section \ref{issue:citation}). %Dataset management and citation facilitates the documentation of datasets, transparency in its use, and the tracking of how datasets are used. 
Dataset documentation is not easily accessible from citations and is not persistent. Moreover, dataset use is not clearly specified in academic papers, often resulting in ambiguities. Finally, current infrastructure does not support the tracking of dataset use or of derivatives in order to retrospectively understand the impact of datasets.
\end{itemize}

%In short, we show how unclear retractions, the creation of derivative datasets and models, broader technological and social change, the lack of clarity of licenses, and dataset management practices can introduce a wide range of ethical concerns.
Based on these findings, we revisit existing recommendations for mitigating the harms that arise from datasets, and adapt them to encompass the broader set of concerns we describe here. Our approach emphasizes steps that can be taken after dataset creation, which we call dataset stewarding. We advocate for responsibility to be distributed among many stakeholders including dataset creators, conference program committees, dataset users, and the broader research community.
\section{Overview of datasets and analysis}
\label{sec:overview}
We first collected a list of 54 face and person recognition datasets (listed in \Cref{app:assembling-datasets}), and chose three popular ones for a detailed analysis of their life cycles: Labeled Faces in the Wild (LFW)~\cite{lfw}, \duke{}~\cite{dukemtmc}, and \msceleb{}~\cite{ms-celeb-1m}. We chose LFW because it was the most cited in our list and allows for longitudinal analysis since it was introduced in 2007.\footnote{LFW had slightly fewer total citations than one other dataset in our list, Yale Face Database B \cite{Georghiades2001FromFT}, but LFW has been cited significantly more times per year, especially in recent years.} We chose \duke{} and \msceleb{} because they were the most cited datasets in our list that had been retracted. We refer to these three datasets as \emph{parent datasets}. We describe them in detail in \Cref{app:dataset-descriptions}.

\begin{table}
\caption{A summary of our overarching analysis of MS-Celeb-1M, DukeMTMC, and LFW.} 
\label{table:analysis-summary}
\vspace*{3mm}
\centering
\footnotesize
\begin{tabular}{lrrr}
\toprule
 & \textbf{MS-Celeb-1M} & \textbf{DukeMTMC} & \textbf{LFW}\\
\midrule
\textbf{Papers that cite dataset or derivatives} & 1,404 & 1,393 & 7,732\\
\textbf{Papers sampled for analysis}
% (20\% of citing papers or 400, whichever is smaller)
& 276 & 275 & 400\\
\textbf{Papers in sample that use dataset or derivatives} & 179 & 114 & 152\\
\textbf{Derivative datasets identified} & 8 & 7 & 20\\
\textbf{Pre-trained models identified} & 21 repositories & 0 & 0\\
\bottomrule
\end{tabular}
\end{table}

We began our analysis by constructing a corpus of papers that cited---and potentially used---each parent dataset or its derivatives (we use the term \emph{derivative} broadly, including datasets that contain the original images, datasets that provide additional annotations, as well as models pre-trained on the dataset). To do this, we first compiled a list of derivatives of each parent dataset and associated them with their research papers. We then compiled a list of papers citing each of these associated papers using the Semantic Scholar API \cite{ssapi}.
% Our corpora for DukeMTMC, MS-Celeb-1M, and LFW contained 1,393, 1,404, and 7,732 papers respectively. Since these were a large number of papers to examine manually, we sampled 20\% or 400 papers (whichever was fewer) stratified over the year of publication.
The first author coded a sample of these papers, recording whether a paper used the parent dataset or a derivative as well as the name of the parent dataset or derivative. In total, our analysis included 946 unique papers, including 275 citing \duke{} or its derivatives, 276 citing \msceleb{} or its derivatives, and 400 citing LFW or its derivatives. We found many papers using derivatives that were not included in our original list of derivatives, which we consider an unavoidable limitation since we are not aware of a systematic way to find all derivatives. Because our corpus does not contain \emph{all} papers using the parent dataset or a derivative, our results should be viewed as lower bounds throughout. We further note that most of our analyses do not address use outside published research. We provide additional details about our methods in \Cref{app:method}.

\section{Retractions and runaway data}\label{issue:decentralized}

When datasets are deemed problematic by the machine learning community, activists, or the media, dataset creators have responded by retracting them. \msceleb{}~\cite{ms-celeb-1m}, \duke{}~\cite{dukemtmc}, VGGFace2~\cite{Cao2018VGGFace2AD}, and Brainwash~\cite{brainwash} were all retracted after an investigation by Harvey and Laplace~\cite{megapixels} highlighted ethical concerns with how the data was collected by the creators and being used by the community. TinyImages~\cite{tinyimages} was retracted after Prabhu and Birhane~\cite{pyrrhic} raised ethical concerns about offensive labels and a lack of consent by data subjects. 

Retractions such as these may mitigate harm in two primary ways. First, they may place hard limitations on dataset use by making the data unavailable. Second, they may exert a normative influence, indicating to the community that the data should no longer be used. This can allow publication venues and other bodies to place their own limitations on such use.

\begin{figure}
    \centering
    \includegraphics[width=\linewidth]{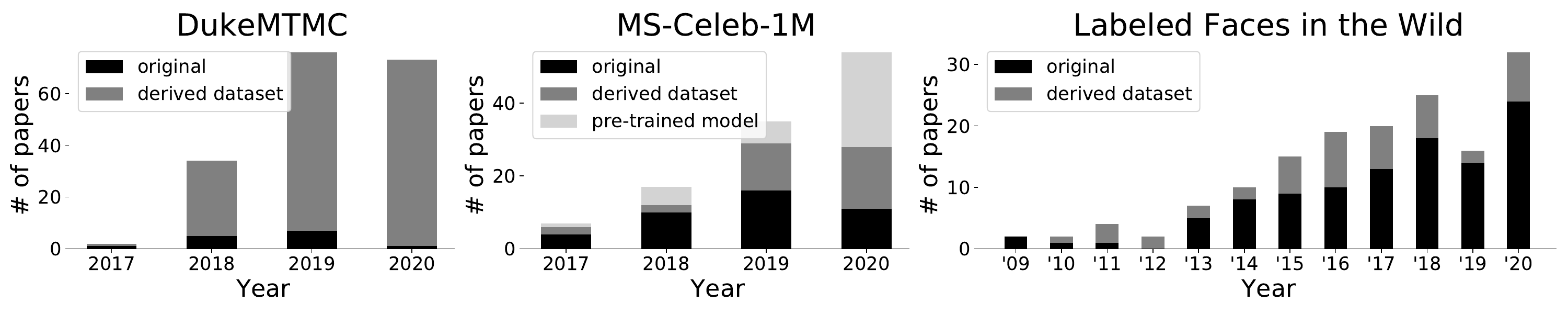}
    \caption{The use of DukeMTMC, MS-Celeb-1M, LFW, and their derivatives over time. All three datasets were commonly used through derivatives. DukeMTMC and MS-Celeb-1M were retracted in April 2019, but continued to be used in 2020---largely, through derivatives.}
    \label{fig:use-over-time}
\end{figure}

With this in mind, we analyzed the retractions of MS-Celeb-1M
\footnote{Although the creators of MS-Celeb-1M never officially stated that the dataset was removed due to ethical concerns, we consider its removal a \emph{retraction} in the sense that its removal responded to ethical concerns. The removal followed mere days after the second of two critical reports of the dataset \cite{megapixels, murgia_2019}. Furthermore, the reason Microsoft gave for the removal---that ``the research challenge is over'' \cite{murgia_2019}---is not, by itself, a common reason to remove a dataset.} 
and DukeMTMC, summarized in Table~\ref{table:retractions}. We find that both retractions fall short of effectively accomplishing either of the above mentioned goals. Since the underlying data was available through many different sources (i.e., the data had ``runaway'' \cite{megapixels}), both datasets remain available despite the retraction of the parent dataset. 
And because the dataset creators did not clearly state that the datasets should no longer be used, they may have left users confused, contributing to their continued use (see \Cref{fig:use-over-time}).

% And because the retractions lacked clarity about whether or not the creators consider it acceptable to use the dataset --- in particular, the websites were simply taken down ---  they left users confused, perhaps contributing to their continued use (see \Cref{fig:use-over-time}).

\begin{table}
\caption{A summary of the status of MS-Celeb-1M and DukeMTMC after their April 2019 retractions.} 
\label{table:retractions}
\vspace*{3mm}
\centering
\small
\begin{tabular}{@{}>{\raggedright}p{2.4cm}p{5.6cm}p{5.3cm}@{}}
\toprule
 & \multicolumn{1}{c}{\textbf{\msceleb{}}} & \multicolumn{1}{c}{\textbf{\duke{}}}\\
\midrule
Availability of original & The dataset is still available through Academic Torrents and Archive.org. & We did not find any locations where the original dataset is still available. \vspace{0.15cm} \\
Availability of derived datasets & We found five derived datasets that remain available with images from the original. & We found two derived datasets that remain available with images from the original. \vspace{0.15cm} \\
Availability of pre-trained models & We found 20 GitHub repositories containing models pre-trained on MS-Celeb-1M that remain available. & We did not find any models pre-trained on DukeMTMC data that are still available. \vspace{0.15cm} \\
Continued use & In our 20\% sample, \msceleb{} and its derivatives were used 54 times in papers published in 2020.\vspace{0.15cm} & In our 20\% sample, \duke{} and its derivatives were used 73 times in papers published in 2020. \vspace{0.15cm} \\
Status of original dataset page & Website (\url{https://www.msceleb.org}) only contains filler text.\vspace{0.15cm} & Website (\url{http://vision.cs.duke.edu/DukeMTMC/}) returns a DNS error. \vspace{0.15cm} \\
Other statements made by creators & In June 2019, Microsoft said in response to a press inquiry that the dataset was taken down ``because the research challenge is over'' \cite{murgia_2019}. & A creator of DukeMTMC apologized in June 2019, noting that they had violated IRB guidelines \cite{tomasi_2019}, but this explanation did not appear in official channels.  \vspace{0.15cm} \\
Availability of metadata & The license is no longer officially available. It was previously available through the website, which was taken down in April 2019. Notably, the license prohibits distribution of the dataset or derivatives. & The license is no longer officially available, but is still available through GitHub repositories of derivative datasets. \\
\bottomrule
\end{tabular}
\end{table}

\new{There are many similarities between the continued use of retracted datasets and the continued citation of retracted papers, which is a well-known yet persistent challenge \cite{Silva2016WhyDS}. Several studies have shown that articles continue to be cited after retraction (e.g., \cite{CandalPedreira2020DoesRA, BarIlan2018TemporalCO, Schneider2020ContinuedPC}). One reason might be because the retraction status is often not clear in all locations where a paper is available \cite{Silva2016WhyDS}. Two primary types of interventions have been proposed to limit continued citation. The first involves making the retraction status of articles more clear and accessible \cite{BornemannCimenti2016PerpetuationOR, oransky_2016}. The second involves publication venues requiring authors to check that their reference list includes no retracted papers \cite{Sox2006ResearchMR, BarIlan2018TemporalCO}. The same types of interventions are applicable in the case of retracted datasets, and are reflected in the recommendations we provide in \Cref{sec:recommendations}.}

In addition to questions of efficacy, retraction can come into tension with efforts to archive datasets.
% Datasets are often seeded on the platforms Academic Torrents and Internet Archive. For example, while the creators TinyImages asked that existing copies of the dataset be deleted, it had already previously been added to both Academic Torrents and Internet Archive. The 2011 version of ImageNet, which contains offensive images that were later removed from official versions, had also previously been added to both sites.
In work critiquing machine learning datasets, \citet{excavatingai} note the issue of “inaccessible or disappearing datasets,” writing that “If they are, or were, being used in systems that play a role in everyday life, it is important to be able to study and understand the worldview they normalize. Developing frameworks within which future researchers can access these data sets in ways that don’t perpetuate harm is a topic for further work.”
\section{Derivatives raise new ethical questions}\label{issue:derivatives}

Machine learning datasets often serve simultaneous roles as a specific tool (e.g., a benchmark for a particular task) and as a collection of raw material that may be leveraged for other purposes. Derivative creation falls into the latter category, and can be seen as a success of resource-sharing in the machine learning community as it reduces the cost of obtaining data. This also means that the effort or cost of creating an ethically-dubious derivative can be much less than creating a similar dataset from scratch. For example, the DukeMTMC-ReID dataset was created using annotations and bounding boxes from the original dataset to build a cropped subset for benchmarking person re-identification. This process can be entirely automated (as far as we can determine from available documentation), which is far cheaper and faster than collecting and manually annotating videos.

In our analysis, we identified four ways in which a derivative can raise ethical considerations (which does not necessarily imply that the creation of the derivative or the parent dataset is unethical). We analyzed all the 41 derivatives of MS-Celeb-1M, DukeMTMC, and LFW based on the four categories we identified. The full matrix is in Table \ref{table:all-use} in the appendix; we summarize the four categories below.

\paragraph{New application.} Either implicitly or explicitly, modifications of a dataset can enable applications raising new ethical concerns. Twenty-one of 41 derivatives we identified fall under this category. For example, DukeMTMC-ReID, a person re-identification benchmark, is used much more frequently than DukeMTMC, a multi-target multi-camera tracking benchmark. While these problems are similar, they may have different motivating applications.
% Several papers flagged by MegaPixels \cite{megapixels} using DukeMTMC data for ethically dubious purposes use DukeMTMC-ReID.
SMFRD \cite{Wang2020MaskedFR} is a derivative of LFW that adds face masks to its images. It is motivated by face recognition applications during the COVID-19 pandemic, when many people wear face-covering masks. “Masked face recognition” has been criticized for violating the privacy of those who may want to conceal their face (e.g., \cite{metz_2020, yan_2021}).

\paragraph{Pre-trained models.}
We found six model classes that were commonly trained on MS-Celeb-1M. Across these six classes, we found 21 GitHub repositories that released models pre-trained on MS-Celeb-1M. These pre-trained models can be used out-of-the-box to perform face recognition or can be used for transfer learning.
% Because the models can already compute salient feature representations of faces, they can be used as the basis for other tasks.
This enables the use of MS-Celeb-1M for a wide range of applications, albeit in a more indirect way.
There are also concerns about the effect of biases in training data on pre-trained models and their downstream applications \cite{Steed2021ImageRL}.

\paragraph{New annotations.}
The annotation of data can also result in privacy and representational harms. (See Section 3.1 of \cite{paullada2020data} for a survey of work discussing representational concerns.) Seven of 41 derivatives fall under this category. Among the derivatives we examined, four annotated the data with gender, three with race or ethnicity, and two with additional attributes such as “attractiveness.” Such annotations may also enable research in ethically dubious applications such as the classification and identification of people via sensitive attributes.

\paragraph{Other post-processing.}
Other derivatives neither repurpose the data for new applications nor contribute annotations. Rather, these derivatives are designed to aid the original task with more subtle modifications. Still, even minor modifications can raise ethical questions.
% For example, Datasheets for Datasets~\cite{datasheets} includes a question about the potential effects of preprocessing or cleaning.\footnote{The question, in full, is: “Is there anything about the composition of the dataset or the way it was collected and preprocessed/cleaned/labeled that might impact future uses?”}
Five of 41 derivatives (each of MS-Celeb-1M) “clean” the original dataset, creating a more accurate set of images from the original, which is known to be noisy. This process often reduces the number of images significantly, after which, we may be interested in the resulting composition. For example, does the cleaning process reduce the number of images of people of a particular demographic group? Such a shift may impact the downstream performance of such a dataset. Five of 41 derivatives (each of LFW) align, crop, or frontalize images in the original dataset. Here, too, we may ask about how such techniques perform on different demographic groups.
\section{Effectiveness of licenses}\label{issue:tos}

Licenses, or terms of use, are legal agreements between the creator and users of datasets, and often dictate how the dataset may be used, derived from, and distributed. We focus on the role of a license in harm mitigation, i.e., as a tool to restrict unintended and potentially harmful uses of a dataset.

By analyzing the licenses of DukeMTMC, MS-Celeb-1M, LFW, and ImageNet, and whether restrictions were inherited by derivatives, we found several shortcomings of licenses as a tool for mitigating harms through preventing commercial use. We included ImageNet in this analysis because we discovered in preliminary research that there is confusion around the implications of ImageNet's license (which allows only non-commercial research use) on pre-trained models. Our findings are summarized in \Cref{table:tos}.

\begin{table}
  \caption{Dataset creators may intend to prohibit commercial use, but licenses do not effectively accomplish this.}
  \label{table:tos}
  \vspace*{3mm}
  \centering
  \small
  \begin{tabular}{p{1.4cm}p{3.6cm}p{3.6cm}p{3.8cm}}
    \toprule
    % \multicolumn{2}{c}{Part}                   \\
    % \cmidrule(r){1-2}
        \textbf{Dataset} & \textbf{Non-commercial intention}  &  \textbf{License shortcomings} & \textbf{Evidence of comm. use}\\
        \midrule
        MS-Celeb-1M & Users may ``use and modify this Corpus for the limited purpose of conducting non-commercial research.'' \vspace{0.15cm}&  Implication on pre-trained models is unclear. The license is no longer publicly available. & We found 18 GitHub repositories containing models pre-trained on MS-Celeb-1M data and released under commercial licenses.\vspace{0.15cm} \\
        LFW & ``... it should not be used to conclude that an algorithm is suitable for any commercial purpose.''\vspace{0.15cm}  &  No license was issued. A disclaimer was added in 2019 (excerpted on left), but carries no legal weight.\vspace{0.15cm} & We identified four commercial systems that actively advertise their performance on LFW.\vspace{0.15cm} \\
        ImageNet & ``Researcher shall use the Database only for non-commercial research and educational purposes.'' & The license does not prevent re-distribution of the data or pre-trained models under commercial licenses.\vspace{0.15cm} & We found nine GitHub repositories containing models pre-trained on ImageNet and released under commercial licenses. Keras, PyTorch, and MXNet include pre-trained weights.\vspace{0.15cm} \\
        DukeMTMC & ``You may not use the material for commercial purposes.''  &  Implication on pre-trained models is unclear. Government use is not ``commercial,'' but can raise similar or greater ethical concerns. & We did not find clear evidence suggesting commercial use of DukeMTMC.\\
    \bottomrule
  \end{tabular}
\end{table}

% \subsection{Commercial use of models trained on non-commercial data}\label{sec:commercial-use}

Motivated by these findings, we further sought to understand whether models trained on datasets released for non-commercial research are being used commercially. Such use can exacerbate the real-world harm caused by datasets. Due to the obvious difficulties involved in studying this question, we approach it by studying online discussions. We identified 14 unique posts on common discussion sites that inquired about the legality of using pre-trained models that were trained on non-commercial datasets. 
% \kenny{REMOVE: This suggests that this is a topic of significant user confusion.}

% \footnote{We identified these posts via four Google searches with the query ``pre-trained model commercial use.'' We then searched the same query on Google with ``site:www.reddit.com,'' ``site:www.github.com,''  ``site:www.twitter.com,'' and ``site:www.stackoverflow.com.'' These are four sites where questions about machine learning are posted. For each search, we examined the top 10 sites presented by Google. Within relevant posts, we also extracted any additional relevant links included in the discussion.}

% Eleven of these posts specifically raised the question of if commercial use of models pre-trained on ImageNet is legal.
% In general, the question of legality is not one we seek to answer; for our purposes, it serves as a window into commercial practices.

From these posts, we found anecdotal evidence that non-commercial dataset licenses are sometimes ignored in practice. One response reads: ``More or less everyone (individuals, companies, etc) operates under the assumption that licences on the use of data do not apply to models trained on that data, because it would be extremely inconvenient if they did.''
% \footnote{\url{https://www.reddit.com/r/MachineLearning/comments/7eor11/d_do_the_weights_trained_from_a_dataset_also_come/}}
Another response reads: ``I don't know how legal it really is, but I'm pretty sure that a lot of people develop algorithms that are based on a pretraining on ImageNet and release/sell the models without caring about legal issues. It's not that easy to prove that a production model has been pretrained on ImageNet ...''
% \footnote{\url{https://www.reddit.com/r/MachineLearning/comments/id4394/d_is_it_legal_to_use_models_pretrained_on/}}
Commonly-used computer vision frameworks like Keras and PyTorch include models pre-trained on ImageNet, making the barrier for commercial use low.

In responses to these posts, representatives of Keras and PyTorch suggested that such use is generally allowed, but that they could not provide an official answer.
% \footnote{\url{https://github.com/Engineering-Course/LIP_JPPNet}}
The representative for PyTorch wrote that according to their legal team's guidance, ``weights of a model trained on that data may be considered derivative enough to be ok for commercial use. Again, this is a subjective matter of comfort. There is no publishable `answer' we can give.''
% \footnote{\url{https://github.com/pytorch/vision/issues/2597}}
The representative for Keras wrote that ``In the general case, pre-trained weight checkpoints have their own license which isn't inherited from the license of the dataset they were trained on. This is not legal advice, and you should consult with a lawyer.''
% \footnote{\url{https://github.com/keras-team/keras/issues/13362}} 
% A representative of LIP\_JPPNet suggested that the user's concern was correct, and that ``You can train the code on your own datasets to get a model for commercial use.''
% \footnote{\url{https://github.com/Engineering-Course/LIP_JPPNet/issues/42}}

While we don't comment on the legality of these practices, we note that they represent a potential legal loophole. If a company were to train a model on ImageNet for commercial purposes, it would be a relatively clear license violation; yet, the practice of downloading pre-trained models, which has substantively the same effect, appears to be common. Similarly, derivatives that don't inherit the license restrictions of the original dataset may also represent a loophole. Dataset creators can avoid such unintended uses by being much more specific in their licenses. For example, The Montreal Data License \cite{montreal} allows for dataset creators to specify restrictions to models trained on the dataset.

\new{We caution that our analysis in this section is preliminary and that the evidence we have presented is tentative and anecdotal. A more thorough study could be conducted through interviews or surveys of practitioners to further illuminate their common practices, legal understanding, as well as the extent to which legal understanding shapes practice.}

\section{Technological and social change affects dataset ethics}\label{issue:context}

We now examine how ethical considerations associated with a dataset change over time. For this analysis, we used LFW and ImageNet, but not DukeMTMC and MS-Celeb-1M as they are relatively recent and thus less fit for longitudinal analysis. We observed that ethical concerns involving both datasets surfaced more than a decade after release. We identified several factors---including increasing production viability, evolving ethical standards, and changing academic incentives---that may help explain this delay.

\begin{figure}
    \centering
    \includegraphics[width=\linewidth]{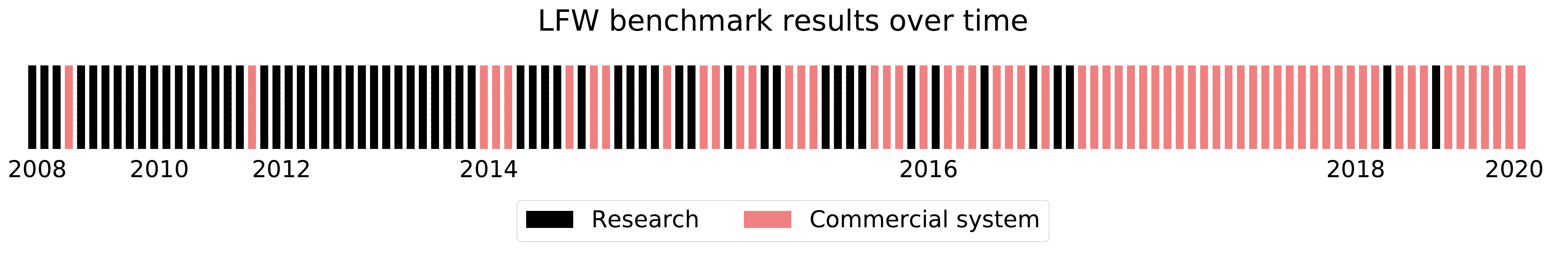}
    \caption{A visualization of the rise of the production use of LFW, based on data from LFW's website. \new{By examining versions of the website archived on the Wayback Machine, we identified (approximately) the year in which different results were added.} Only 3 of 38 results added before 2014 were commercial but 41 of 49 results after 2016 were  commercial.}
    \label{fig:lfw-benchmark}
\end{figure}

\paragraph{Changing ethics of LFW.} LFW was introduced in 2007 to benchmark face verification. It is considered the first “in-the-wild” face recognition benchmark, designed to help face recognition improve in  unconstrained settings. The production use of the dataset was unviable in its early years, one indication being that the benchmark performance on the dataset was poor.\footnote{Consider the benchmark task of face verification with unrestricted outside training data (the easiest of the tasks proposed in \cite{lfw}). The best reported accuracy as of 2010 was only  84.5\%, whereas today it is 99.9\%.} Over time, LFW became a standard benchmark and technology improved. This opened to door for increased use of LFW to benchmark commercial systems, as illustrated in Figure~\ref{fig:lfw-benchmark}. 

This type of use inspired ethical concerns, as benchmarking production systems has greater real-world potential for harm than benchmarking models used for research. The production use of facial recognition systems in applications such as surveillance or policing have caused backlash---especially because of disparate performance on minority groups.

In 2019---more than a decade after the dataset's release---a disclaimer was added to the dataset's website noting that it should not be used to verify the performance of commercial systems \cite{lfw-web}. Notably, this disclaimer emphasized LFW's insufficient diversity across many demographic groups, as well as in pose, lighting, occlusion, and resolution. In contrast, when the dataset was first released, the creators highlighted the dataset's diversity: LFW contained real-world images of people, whereas past datasets had mostly contained images taken in a laboratory setting \cite{aboutface}. This shift may be partially due to recent work showing disparate performance of face recognition  on different demographic groups and highlighting the need for demographically-diverse benchmarks \cite{Buolamwini2018GenderSI}.

\paragraph{Changing ethics of ImageNet.} When ImageNet was introduced, object classification was still immature. Today, as real-world use of such technology has become widespread, ImageNet has become a common source for pre-training, again illustrating the shift from research to production use. As discussed in Section~\ref{issue:tos}, even as the dataset’s terms of service specify non-commercial use, the dataset is commonly used in pre-trained models released under commercial licenses. %Weights pre-trained on ImageNet come built-in in popular machine learning frameworks such as Keras and PyTorch.

We also consider how social factors have shaped recent ethical concerns. In 2019, researchers revealed that many of the images in the “people” category of the dataset were labeled with misogynistic and racial slurs and perpetuated stereotypes, after which images in these categories were removed~\cite{excavatingai, pyrrhic}. This work critiquing ImageNet first appeared nearly a decade after its release (even if issues were known to some earlier). As it is reasonable to assume that the labels used in ImageNet would have been considered offensive in 2009, the lag between the dataset's release and the removal of such labels is noteworthy. We propose three factors that have changed since the release of ImageNet and hypothesize that they may account for the lag. First, public concern over machine learning datasets and applications has grown. Issues involving datasets have received significant public attention---the article by Crawford and Paglen \cite{excavatingai} accompanied several art exhibitions and the topic has been covered by many media outlets (e.g., \cite{murgia_2019, metz_2019, solon_19AD}). Relatedly, academic incentives have changed and critical work is more easily publishable. Related work highlighting assumptions underlying classification schemes \cite{Benthall2019RacialCI, Khan2021OneLO} have been published in FAccT, a conference focused on fairness, accountability, and transparency in socio-technical systems that was only founded in 2018. Finally, norms regarding the ethical responsibility of dataset creators and machine learning researchers more generally have shifted. These norms are still evolving; responses to recently-introduced ethics-related components of peer review have been mixed \cite{abuhamad2020like}.

\paragraph{} The transition from research to production use, in some sense, is a sign of success of the dataset, and thus may be anticipated. Benchmark datasets in machine learning are typically introduced for problems that are not yet viable in production use cases; and should the benchmark be successful, it will help lead to the realization of real-world application. \new{The ethics of LFW and ImageNet were also each shaped by social factors, if in different ways. Whereas shifting ethical standards contributed to changing views of LFW, ImageNet labels would likely have been considered offensive when the dataset was first created. For ImageNet, social factors seem to have led to evolving incentives to identify and address ethical issues. While ``what society deems fair and ethical changes over time'' \cite{Birhane2019AlgorithmicIT}, additional factors can dictate if and how these standards are operationalized.}

\section{Dataset management and citation}\label{issue:citation}

We turn to the role of dataset management and citation in harm mitigation. By dataset management, we mean storing a dataset and associated metadata. By dataset citation, we mean the referencing of a dataset used in research with the aim of facilitating access to the dataset and metadata. 
% Though our focus is on harm mitigation, dataset management and citation also help scientific communities function more efficiently in general \cite{Wilkinson2016TheFG}. 
We give three reasons for why dataset management and citation are important for mitigating harms caused by datasets: facilitating documentation accessibility, transparency and accountability, and tracking of dataset use. We then summarize how current practices fall short in achieving these aims.

% \subsection{Dataset management and citation can help mitigate harms}

% We give three reasons for why dataset management and citation are important for mitigating harms caused by datasets.

\paragraph{Documentation.} Access to dataset documentation facilitates responsible dataset use. Documentation can provide information about a dataset’s composition, its intended use, and any restrictions on its use (through licensing information, for example). Many researchers have proposed documentation tools for machine learning datasets with harm mitigation in mind \cite{datasheets, nlp-data-statements}. Dataset management and citation can ensure that documentation is easily accessible, even if the dataset itself is not or is no longer publicly accessible. In Section~\ref{issue:decentralized} and Section~\ref{issue:tos}, we discussed how retracted datasets no longer included key information such as licensing information, potentially leading to confusion. For example, with MS-Celeb-1M’s license no longer publicly available, the license status of derivative datasets, pre-trained models, and remaining copies of the original is unclear.

\paragraph{Transparency and accountability.} Dataset citation facilitates transparency in dataset use, in turn facilitating accountability. By clearly indicating the dataset used and where information about the dataset can be found, researchers become accountable for ensuring the quality of the data and its proper use. Different stakeholders, such as the dataset creator, program committees, and other actors can then hold researchers accountable. For example, if proper citation practices are followed, peer reviewers can more easily check whether researchers complied with dataset licenses.

\paragraph{Tracking.} 
Large-scale analysis of dataset use---as we do in this paper---can illuminate a dataset’s impact and potential avenues of risk or misuse. This knowledge can allow dataset creators to update documentation, better establishing intended use. Citation infrastructure supports this task by collecting such use in an organized manner. This includes both tracking the direct use of a dataset in academic research, as well as the creation of derivatives.

Our findings, summarized below, suggest that current dataset management and citation practices fall short in supporting the above goals. A complete set of findings is given in \Cref{app:citation}.

\begin{itemize}[leftmargin=*]
\item \textbf{Datasets and metadata are not persistent.} None of the 38 datasets in our analysis are managed through shared repositories, a common practice in other scientific fields. We were unable to locate three datasets, and two more are only available through the Wayback Machine. After DukeMTMC and MS-Celeb-1M's retractions, their licenses are no longer officially available.
\item \textbf{Disambiguating citations is hard.} None of the 38 datasets have DOIs or stable identifiers. Only six of 60 sampled papers provided access information such as a URL. We encountered difficulties accessing datasets when no URL is given, as five of the datasets did not even have names. The current practice of citing datasets via a combination of name, description, and associated papers makes even manual disambiguation challenging. We were unable to disambiguate a citation in 42 of 446 cases and encountered difficulties in roughly 50 additional cases.
\item \textbf{Tracking is difficult.} The lack of dataset-specific identifiers makes systematic tracking hard. Papers using a dataset may not cite a particular paper, and vice versa. Moreover, there is no way to systematically identify the derivatives of a dataset.
\end{itemize}

\section{Recommendations}\label{sec:recommendations}

%\anote{TODO: get rid of almost everything before the recommendations with a few general observations: (1) there is no silver bullet; (2) ethical responsibilities must be distributed among stakeholders (3) ethical responsibilities continue throughout the lifecycle (4) dataset creation must be incentivized (5) none of this is of much help if people are malicious (6) our recommendations are not mutually exclusive with other suggestions such as datasheets}

In the last few years, there have been numerous recommendations for mitigating the harms associated with machine learning datasets. \new{Researchers have proposed frameworks for dataset and model documentation \cite{datasheets, nlp-data-statements, Holland2018TheDN}, which can both guide responsible dataset creation and facilitate responsible use. Other researchers have proposed guidelines for ethical data collection \cite{ethicalhighlighter}, drawing from ``interventionist'' approaches modeled on archives \cite{Jo2020LessonsFA} or focusing on specific principles, such as requiring informed consent from data subjects \cite{pyrrhic}. Still others have created tools for identifying and mitigating biases \cite{aifairness360, revise} or preserving privacy \cite{pyrrhic, Yang2021ASO} in datasets.} Our own recommendations build on this body of work and aren't meant to replace existing proposals. 

That said, previous approaches primarily consider dataset creation. As we have shown, ethical impacts are hard to anticipate and address at dataset creation time. Thus, we argue that harm mitigation requires stewarding throughout a dataset's life cycle. Our recommendations reflect this understanding. 

We contend that the problem cannot be left to any one stakeholder such as dataset creators or IRBs. We propose a more distributed approach where many stakeholders share responsibility for ensuring the ethical use of datasets. We assume the willingness of dataset creators, program committees, and the broader research community; addressing harms from callous or malicious users or outside the research context is beyond our scope. Below, we present recommendations for dataset creators, conference program committees, dataset users, and other researchers. In \Cref{app:recommendations}, we discuss how IRBs---which hold traditional oversight over research ethics---are an imperfect fit for dataset-centered research and should not be relied on for regulating machine learning datasets or their use.

\new{Our recommendations are informed by the principle of separating blame from responsibility. Even if an entity is not to blame for a particular harm, that entity might be well positioned to reduce the likelihood of that harm occurring. For example, as a response to ML research that develops technologies that could be used used to violate human rights, it is reasonable to allocate some responsibility to conference program committees to prohibit this type of research. Similarly, as a response to harms associated with data, it is reasonable to allocate some responsibility to dataset creators. As we argue below, there are many ways in which dataset creators can minimize the chances of downstream abuse.}

\subsection{Dataset creators}\label{rec:creators}

We make two main recommendations for dataset creators, both based on the normative influence they can exert and based on the harder constraints they can impose on how datasets are used. 

\paragraph{Make ethically salient information clear and accessible.} Dataset creators can place restrictions on dataset use through licenses and provide other ethically salient information through other documentation. But in order for these restrictions to be effective, they must be clear. In our analysis, we found that licenses are often insufficiently specific. For example, when restricting the use of a dataset to ``non-commercial research,'' creators should be explicit about whether this also applies to models trained on the dataset. It may also be helpful to explicitly prohibit specific ethically questionable uses.
% The Casual Conversations dataset does this  (``Participants will not ... use the Casual Conversations Dataset to measure, detect, predict, or otherwise label the race, ethnicity, age, or gender of individuals, label facial attributes of individuals, or otherwise perform biometric processing unrelated to the Purpose'') \cite{casual}. The Montreal Data License \cite{montreal} is an example of a license that allows for specific designations. Going beyond licenses, Datasheets for Datasets provides a standard for detailed and clear documentation of ethically salient information \cite{datasheets}.

In order for licenses or documentation to be effective, they also need to be accessible. Licenses and documentation should be persistent, which can be accomplished through the use of standard data repositories. Dataset creators should also set requirements for dataset users and creators of derivatives to ensure that this information is easy to find from citing papers and derived datasets.

\new{These recommendations also apply to dataset retraction. Retractions should be explicit and easily accessible. Moreover, dataset creators should seek to make the retraction status visible wherever the dataset or its derivatives remain available.}

\paragraph{Actively steward the dataset and exert control over use.}

Throughout our analysis, we show how ethical considerations can evolve over time. Dataset creators should continuously steward a dataset, actively examining how it may be misused, and making updates to the license, documentation, or access restrictions as necessary. A minimal access restriction is for users to agree to terms of use. A more heavyweight process in which dataset creators make case-by-case decisions about access requests can be used in cases of greater risk. The Fragile Families Challenge is an example of this \cite{fragile-families}. 

Based on our analysis in Section~\ref{issue:decentralized} and  Section~\ref{issue:derivatives}, derivative creation often raises ethical risks. \new{We showed that derivatives can make data more widely available---in many cases, without the original licensing information. Additionally, derivatives may introduce new ethical concerns, such as through enabling new applications. A dataset's terms of use can establish guidance for derivative creation. This may include a list of specifically allowed (or disallowed) types of derivatives, in addition to distribution and licensing requirements. Of course, dataset creators may not be able to anticipate all potential ethically-dubious derivatives in advance. Creators may overcome this by requiring explicit permission be obtained unless the derivative belongs to a pre-approved category.}

\new{We recognize that dataset stewarding increases the burden on dataset creators. In our discussions with dataset creators, we heard that  creating datasets is already an undervalued activity and that a norm of dataset stewardship might further erode the incentives for creators. We acknowledge this concern, but maintain that there is an inherent tension between ethical responsibility and  minimizing burdens on dataset creators. One solution is for dataset creation to be better rewarded in the research community; some of our suggestions for program committees below may have this effect. }

\subsection{Conference program committees}\label{rec:pcs}

%In the machine learning community and the broader computer science community, the primary publication venues are the peer-reviewed proceedings of conferences, and the prestige associated with these publications is the core of the reward structure for researchers. Thus, conference program committees (PCs) hold immense power to set ethical standards for the research community including both dataset creators and users. Of course, PCs have little power over commercial use of datasets that is not motivated by publications. Ethics review as part of peer review is a recent development at machine learning conferences but has a longer history in the computer security community.

\paragraph{Use ethics review to encourage responsible dataset use.}

PCs are in a position to govern both the creation of datasets (and derivatives) and the use of datasets through ethics reviews of the associated papers. 
PCs should develop clear guidelines for ethics review. For example, PCs can require researchers to clearly state the datasets used, justify the reasons for choosing those datasets, and certify that they complied with the terms of use of each dataset. \new{In particular, PCs can require researchers to examine if a dataset has been retracted.} Some conferences, such as NeurIPS, already have ethics guidelines relating to dataset use.

\paragraph{Encourage standard dataset management and citation practices.}

PCs should consider standardized dataset management and citation requirements, such as requiring dataset creators to upload their dataset and supporting documentation to a public data repository. Guidelines on effective dataset management and citation practices can be found in \cite{Wilkinson2016TheFG}. The role of PCs is particularly important for dataset management and citation, as these practices benefit from community-wide adoption. 

\paragraph{Introduce a dataset-specific track.} NeurIPS now includes a track specifically for datasets. The introduction of such tracks facilitates more careful and tailored ethics reviews for datasets. The journal \emph{Scientific Data} is devoted entirely to describing datasets.

\paragraph {Allow advance review of datasets and publications.}
We tentatively recommend that conferences can allow researchers to submit proposals for datasets prior to creation. By receiving preliminary feedback, researchers can be more confident that their dataset both will be valued and will pass initial ethics reviews. This mirrors the concept of ``registered reports'' \cite{registered_reports}, in which a proposed study is peer reviewed before it is undertaken and provisionally accepted for publication {\em before} the outcomes are known, as a way to counter publication biases. 
%As a similar example, the journal JQD:DM requires a detailed letter of inquiry before submitting a paper \cite{jqddm}. 

\subsection{Dataset users and other researchers}\label{rec:users}

At a minimum, dataset users should comply with the terms of use of datasets. But their responsibility goes beyond compliance. They should also carefully study the accompanying documentation and analyze the appropriateness of using the dataset for their particular application (e.g., whether dataset biases may propagate to models). Dataset users should also clearly indicate what dataset is being used in their research papers and ensure that readers can access the dataset based on the information provided. As we showed in Section~\ref{issue:citation}, traditional paper citations often lead to ambiguity. 

% TODO: expand, mention that some dataset creators may have dropped the ball

We showed how a dataset’s impact is not fully understood at the time of its creation. We recommend that the community systematize the retrospective study of datasets to understand their shortcomings and misuse. Researchers should not wait until the problems become serious and there is backlash. 

It is especially important to understand how datasets and pre-trained models are being used in production settings, which our work does not address. Policy interventions may be necessary to incentivize companies to be transparent about  datasets or models used in deployed pipelines. %Policy makers should consider legal requirements that encourage more transparency around the specifics of training datasets used in commercially deployed models.

\section{Conclusion}

The machine learning community is responding to a wide range of ethical concerns regarding datasets and asking fundamental questions about the role of datasets in machine learning research. In this paper, we provided a new perspective. Through our analysis of the life cycles of three datasets, we showed how developments that occur after dataset creation can impact the ethical consequences, making them hard to anticipate a priori. We advocated for an approach to harm mitigation in which responsibility is distributed among stakeholders and continues throughout the life cycle of a dataset.

\section*{Acknowledgments}
This work is supported by the National Science Foundation under Awards IIS-1763642 and CHS-1704444. We thank participants of the Responsible Computer Vision (RCV) workshop at CVPR 2021, the Princeton Bias-in-AI reading group, and the Princeton CITP Work-in-Progress reading group for useful discussion. We also thank Solon Barocas, Marshini Chetty, Sayash Kapoor, Mihir Kshirsagar, and Olga Russakovsky for their helpful feedback.

{
\small

\begingroup
\raggedright
\bibliography{references}
\bibliographystyle{plainnat}
\endgroup
}

\newpage
\appendix
\addcontentsline{toc}{section}{Appendix} % Add the appendix text to the document TOC
\part{Appendix} % Start the appendix part
\parttoc % Insert the appendix TOC
\section{Broader impacts}\label{app:broader-impacts}
The authors discussed potential negative impacts among ourselves. The primary potential negative impact we identified is that we raise awareness of datasets and pre-trained models derived from retracted datasets or released under licenses not adherent to the intent of the license of the parent. However, many of these datasets and pre-trained models are already widely used and accessible, so we do not believe our documentation will cause much additional harm. Instead, we hope that our work makes ethical considerations more clear to users of these assets.

\section{Supplemental data}\label{app:assembling-datasets}
Supplemental data is available at \url{https://github.com/citp/mitigating-dataset-harms}.

\paragraph{Maintenance.} \new{This data will remain available indefinitely as long as the Princeton CITP GitHub is operational. Links used to access publicly-available PDFs may eventually deprecate, but the DOIs we give ensure that all papers included in our analysis remain identifiable.}

\paragraph{License.} \new{License details for our data can be found at the above link: \url{https://github.com/citp/mitigating-dataset-harms}.}

\paragraph{Available files.} We make the following \texttt{.csv} files available:

\paragraph{\texttt{msceleb1m\_all.csv, dukemtmc\_all.csv, lfw\_all.csv}} These are the full corpora we collected, containing 1,404, 1,393, and 7,732 papers respectively. The following columns are given, and reflect information given by Semantic Scholar:
\begin{itemize}[leftmargin=*]
    \item \textbf{paperId:} the Semantic Scholar id of the paper
    \item \textbf{cites \{dataset id\}:} for each dataset used to build the corpus, 1 if the paper cites \{dataset id\} and 0 otherwise---see \Cref{table:all-use} for dataset ids.
    \item \textbf{title, abstract, year, venue, arxivId, doi}
    \item \textbf{pdfUrl:} a URL where the paper may be publicly available, found via Semantic Scholar or arXiv
\end{itemize}

\paragraph{\texttt{msceleb1m\_labeled.csv, dukemtmc\_labeled.csv, lfw\_labeled.csv}} These are the samples of papers that we analyzed, containing 276, 275, and 400 papers respectively. In addition to all the columns above, the following additional columns are given:
\begin{itemize}[leftmargin=*]
    \item \textbf{uses dataset or derivative:} 1 if we determined that the paper uses a dataset or derivative and 0 otherwise
    \item \textbf{dataset(s) / model(s) used:} a comma separated list of datasets or models used, denoted by the id provided in \Cref{table:all-use} in brackets (e.g., [D8], [M5])
    \item \textbf{unable to disambiguate:} 1 if we were unable to determine the specific dataset(s) used or whether a dataset was used, and 0 otherwise
\end{itemize}

\paragraph{\texttt{summary.csv}} An extended version of \Cref{table:all-use}.

\paragraph{\texttt{dataset\_list.csv}} This file contains the names of the 54 face and person recognition datasets we compiled to select our three datasets of interest, the number of total citations on Semantic Scholar (at the time of collection in August 2020), and their Semantic Scholar Corpus ID which can be used to access metadata from the Semantic Scholar API \cite{ssapi}.

This list was compiled from datasets mentioned in the following six sources \cite{Cao2018VGGFace2AD, Zheng2020ASO, Li2018DeepFE, Wang2021DeepFR, megapixels, Gross2008MultiPIE} and cited at least 100 times. Sorted from most total citations to least, the datasets are:
Yale Face Database B, LFW, FERRET, AR, VGGFace, LFWA, CelebA, AU-Coded Facial Expression Database (Cohn-Kanade), FRGCv2, Extended Yale Face Database B, CK+, JAFFE, PIE, Multi-PIE, RaFD, PubFig, CelebFaces+, XM2VTS, BU-3DFE, CASIA-WebFace, YTF, MMI, MORPH, DukeMTMC, MS-Celeb-1M, Bosphorus, VGGFace2, CUFS, Replay-Attack, IJB-A, YTC, Attributes 25K, MegaFace, FER-2013, FaceTracer, CASIA-FASD, Berkeley Human Attributes, Oulu-CASIA, AffectNet, Brainwash, CACD, EmotioNet, SPEW 2.0, PaSC, CUFSF, CFP, CASIA NIR-VIS v2.0, RAF-DB, IJB-C, IJB-B, Oxford TownCentre, CASIA-HFB, UMDFaces, and Ego-Humans.

\section{Description of datasets analyzed}\label{app:dataset-descriptions}

In this section, we provide details about the three datasets our analysis focused on: MS-Celeb-1M, DukeMTMC, and Labeled Faces in the Wild.

\textbf{\msceleb{}} was introduced by Microsoft researchers in 2016 as a face recognition dataset~\cite{ms-celeb-1m}. It includes about 10 million images of about 10,000 ``celebrities.'' The original paper gave no specific motivating applications, but did note that ``Recognizing celebrities, rather than a pre-selected private group of people, represents public interest and could be directly applied to a wide range of real scenarios.'' Researchers and journalists noted in 2019 that many of the ``celebrities'' were in fact fairly ordinary citizens, and that the images were aggregated without consent \cite{megapixels, murgia_2019}. Several corporations tied to mass surveillance operations were also found to use the dataset in research papers \cite{megapixels, murgia_2019}. The dataset was taken down in April 2019. Microsoft, in a statement to the Financial Times, said that the reason was ``because the research challenge is over.''~\cite{murgia_2019}

\textbf{\duke{}} was introduced in 2016 as a benchmark for evaluating multi-target multi-camera tracking systems, which ``automatically track multiple people through a network of cameras.''~\cite{dukemtmc} The dataset's creators defined performance measures aimed at applications where preserving identity is important, such as ``sports, security, or surveillance.'' The images were collected from video footage taken on Duke University's campus. The same reports on \msceleb{} listed above \cite{megapixels, murgia_2019} noted that the \duke{} was also being used by corporations tied to mass surveillance operations, and also noted the lack of consent given by people included in the dataset. The creators removed the dataset in April 2019, and subsequently apologized, noting that they had inadvertently broken guidelines provided by the Duke University IRB.

\textbf{LFW} was introduced in 2007 as a benchmark dataset for face verification~\cite{lfw}. It was one of the first face recognition datasets that included faces from an unconstrained ``in-the-wild'' setting, using faces scraped from Yahoo News articles (via the Faces in the Wild dataset \cite{Berg2004WhosIT}). In the originally-released paper, the dataset's creators gave no motivating applications or intended uses beyond studying face recognition. In fall 2019, a disclaimer was added to the dataset's associated website, noting that the dataset should not be used to ``conclude that an algorithm is suitable for any commercial purpose.''~\cite{lfw-web}

% ------------------------------------
% ~~~NEURIPS VERSION OF TABLE BELOW~~~
% ------------------------------------
% ------------------------------------

\begin{table}
  \caption{Summary of our overarching analysis.}
  \vspace*{3mm}
  \centering
  \footnotesize
\begin{tabular}{p{0.3cm}p{3.2cm}p{0.2cm}p{1cm}p{0.2cm}p{0.3cm}p{0.3cm}p{0.2cm}p{0.2cm}p{0.2cm}p{0.2cm}p{0.2cm}p{0.2cm}p{1cm}}
\rot{Dataset id}&\rot{Dataset name}&\rot{Associated paper}&\rot{Dataset or model}&\rot{Assoc. paper sampled}&\rot{Num. sampled}&\rot{Doc. uses}&\rot{2020 doc. uses}&\rot{New application}&\rot{Attribute annotations}&\rot{Post-processing}&\rot{Still available}&\rot{Includes orig. imgs.}&\rot{Prohibits comm. use}\\ 
\midrule
D1& DukeMTMC&\cite{dukemtmc}&dataset&\checkmark&164&14&1&&&&&\checkmark&\checkmark \\
D2& DukeMTMC-ReID&\cite{dukemtmc-reid}&dataset&\checkmark&172&142&63&\checkmark&&&\checkmark&\checkmark&\checkmark \\
D3& DukeMTMC-VideoReID&\cite{Wu2018ExploitTU}&dataset&\checkmark&24&11&5&\checkmark&&&\checkmark&\checkmark&\checkmark \\
D4& DukeMTMC-Attribute&\cite{Lin2019ImprovingPR}&dataset&&&10&1&\checkmark&\checkmark&&\checkmark&&\checkmark \\
D5& DukeMTMC4ReID&\cite{Gou2017DukeMTMC4ReIDAL}&dataset&&&3&0&\checkmark&&&&\checkmark&\checkmark\\
D6*& DukeMTMC Group&\cite{Xiao2018GroupRL}&dataset&&&3&1&\checkmark&&&&&\\
D7& DukeMTMC-SI-Tracklet&\cite{Li2020UnsupervisedTP}&dataset&&&1&1&\checkmark&&&&\checkmark&\checkmark\\
D8& Occluded-DukeMTMC&\cite{Miao2019PoseGuidedFA}&dataset&&&1&1&\checkmark&&&\checkmark\tablefootnote{The dataset itself is no longer available. However, a script to convert DukeMTMC-ReID (which is still available) to Occluded-DukeMTMC remains available.}&\checkmark&\checkmark\\
\midrule
D9& MS-Celeb-1M&\cite{ms-celeb-1m}&dataset&\checkmark&153&41&11&&&&\checkmark&\checkmark& \checkmark\\
D10& MS1MV2&\cite{Deng2019ArcFaceAA}&dataset&\checkmark&183&13&8&&&\checkmark&\checkmark&\checkmark&\\
D11& MS1M-RetinaFace&\cite{Deng2019LightweightFR}&dataset&&&2&2&&&\checkmark&\checkmark&\checkmark&\\
D12& MS1M-LightCNN&\cite{wu2018light}&dataset&&&3&0&&&\checkmark&\checkmark&&\\
D13& MS1M-IBUG&\cite{Deng2017MarginalLF}&dataset&&&3&1&&&\checkmark&\checkmark&\checkmark&\\
D14& MS-Celeb-1M-v1c&\cite{trillionpairs}&dataset&&&6&4&&&\checkmark&\checkmark&\checkmark&\checkmark\\
D15& RFW&\cite{rfw}&dataset&&&1&1&\checkmark&\checkmark&&\checkmark&\checkmark&\checkmark\\
D16& MS-Celeb-1M lowshot&\cite{ms-celeb-1m}&dataset&&&4&0&&&&&\checkmark&\checkmark\\
D17*& Universe&\cite{Bansal2018DeepFF}&dataset&&&2&1&&&&&&\\
M1 & VGGFace&&model&&&6&3&\checkmark&&&\checkmark&&\\
M2 & Prob. Face Embeddings&&model&&&1&1&\checkmark&&&\checkmark&&\\
M3 & ArcFace / InsightFace&&model&&&14&13&\checkmark&&&\checkmark&&some\\
M4 & LightCNN&&model&&&4&3&\checkmark&&&\checkmark&&some\\
M5 & FaceNet&&model&&&12&5&\checkmark&&&\checkmark&&\\
M6 & DREAM&&model&&&1&1&\checkmark&&&\checkmark&&\\
\midrule
D18& LFW&\cite{lfw}&dataset&\checkmark&220&105&&&&&\checkmark&\checkmark&\\
D19& LFWA&\cite{lfwa}&dataset&\checkmark&158&2&&\checkmark&\checkmark&&\checkmark&\checkmark&\\
D20& LFW-a&\cite{lfw-a}&dataset&\checkmark&31&14&&&&\checkmark&\checkmark&\checkmark&\\
D21& LFW3D&\cite{Hassner2015EffectiveFF}&dataset&\checkmark&24&3&&&&\checkmark&\checkmark&\checkmark&\\
D22& LFW deep funneled&\cite{Huang2012LearningTA}&dataset&\checkmark&18&4&&&&\checkmark&\checkmark&\checkmark&\\
D23& LFW crop&\cite{Sanderson2009MultiRegionPH}&dataset&\checkmark&8&2&&&&\checkmark&\checkmark&\checkmark&\\
D24& BLUFR protocol&\cite{Liao2014ABS}&dataset&\checkmark&2&1&&&&&\checkmark&&\\
D25*& LFW87&\cite{Liang2008FaceAV}&dataset&\checkmark&7&1&&&&&&&\\
D26& LFW+&\cite{Han2018HeterogeneousFA}&dataset&\checkmark&12&0&&\checkmark&\checkmark&&\checkmark&\checkmark&\checkmark\\
D27& <no name given> &\cite{kumar}&dataset&&&4&&\checkmark&\checkmark&&\checkmark&&\\
D28& <no name given>&\cite{Guillaumin2009IsTY}&dataset&&&4&&&&&\checkmark&&\\
D29& SMFRD&\cite{Wang2020MaskedFR}&dataset&&&1&&\checkmark&&&\checkmark&\checkmark&\\
D30& LFW funneled&\cite{Huang2007a}&dataset&&&2&&&&\checkmark&\checkmark&\checkmark&\\
D31& <no name given> &\cite{fipa}&dataset&&&2&&\checkmark&\checkmark&&&&\\
D32& <no name given>&\cite{BestRowden2014UnconstrainedFR}&dataset&&&1&&&&&\checkmark&&\\
D33& MTFL&\cite{Zhang2014FacialLD}&dataset&&&1&&\checkmark&\checkmark&&\checkmark&\checkmark&\\
D34& PubFig83 + LFW&\cite{Becker2013EvaluatingOF}&dataset&&&2&&&&&\checkmark&\checkmark&\\
D35& Front. Faces in the Wild&\cite{Ferrari2016Effective3B}&dataset&&&1&&&&&\checkmark&\checkmark&\\
D36& ITWE&\cite{Zhou2017DeformableMO}&dataset&&&1&&\checkmark&&&\checkmark&\checkmark&\checkmark\\
D37& Extended LFW&\cite{Sun2013DeepCN}&dataset&&&2&&&&&\checkmark&\checkmark&\\
D38& <no name given>&\cite{Dantone2012RealtimeFF}&dataset&&&1&&&&&&&\checkmark\\
\bottomrule
\end{tabular}

\caption*{\footnotesize \textbf{Condensed key for Table~\ref{table:all-use}.} \textit{assoc. paper sampled} — yes if our corpus included a sample of papers citing the dataset's associated paper(s); \textit{doc. uses} --- the number of uses of the dataset that we were able to document;  \textit{new application} — if the derivative explicitly or implicitly enables a new application that can raise ethical questions; \textit{attribute annotation} — if the derivative includes labels for sensitive attributes such as race or gender; \textit{post-processing} — if the derivative manipulates the original images (for example, by cleaning or aligning); \textit{prohibits comm. use} — if the dataset or model's license information includes a non-commercial clause; in \textit{dataset id}, an asterisk (*) indicates that we were unable to identify where the dataset is or was made available; in \textit{dataset name}, some datasets were not given names by their creators.}
  \label{table:all-use}
\end{table}

\section{Methodology for overarching analysis}\label{app:method}

We started our analysis of \duke{}, \msceleb{}, and LFW by using the Semantic Scholar API~\cite{ssapi} to record all papers citing their associated papers. Because papers that used derivatives may not always cite the original dataset, we also aimed to pull papers citing associated papers of derivatives. We identified these in a semi-automated fashion: From the list of papers above, we first downloaded PDF versions when they were publicly available either through arXiv or via links provided by Semantic Scholar. We used GROBID \cite{GROBID} to parse these PDFs into plaintext. We then pulled short excerpts containing keywords related to the parent dataset, which we identified through a preliminary review of papers using the dataset. By manually analyzing these excerpts, we identified derivatives that contained these keywords. We further analyzed a sample of papers to identify additional derivatives that did not contain these keywords. We retained derivatives that were cited at least 100 times to build a corpus of papers.

We combined the three parents datasets with these compiled derivatives, and recorded all the papers that cited these datasets, again using the Semantic Scholar API. The resulting corpora for DukeMTMC, MS-Celeb-1M, and LFW contained 1,393, 1,404, and 7,732 papers respectively. These corpora—assembled in December 2020—contain a large subset of papers using the three parent datasets and their derivatives. However, the corpora do not include all papers using the parent datasets and their derivatives. There are a few reasons for this. Our corpora only includes papers added to Semantic Scholar by December 2020, and Semantic Scholar itself does not index all papers.\footnote{We reproduced our corpora in August 2021, and found small discrepancies compared to our corpora collected in December 2020. The number of 2020 papers in our initial corpora for MS-Celeb-1M, DukeMTMC, and LFW are 8\% less, 3\% less, and 1\% more, respectively, compared to the August 2021 version.} Some papers are also missing because the list of derivatives we used to build each corpus is not complete. This means that the results presented throughout our paper are underestimates.

Since these were a large number of papers to examine manually, we sampled 20\% or 400 papers (whichever was fewer) stratified over the year of publication.\footnote{We did not consider years with fewer than 10 papers. We only considered academic papers (both preprint and publications), ignoring other articles like dissertations and textbooks.} In total, our analysis included 946 unique papers: 275 citing \duke{} or its derivatives, 276 citing \msceleb{} or its derivatives, and 400 citing LFW or its derivatives. The first author coded these papers, recording whether a paper used the parent dataset or a derivative as well as the name of the parent dataset or derivative. If the first author was unable to determine the specific dataset used or whether a dataset was used, he recorded this information. A few example cases that were difficult to disambiguate are shown in \Cref{table:citation-disambiguation}.

A summary of our overarching analysis is given in \Cref{table:all-use}.

\section{Supplement: Retractions of DukeMTMC and MS-Celeb-1M}

We describe in detail our findings summarized in \Cref{table:retractions} about the retractions of MS-Celeb-1M and DukeMTMC.

\paragraph{Continued availability.} Despite their retractions in April 2019, data from \msceleb{} and \duke{} remain publicly available. Five of the seven derivatives of \duke{} either contained subsets of or the entire original dataset. The two most popular derivatives---\duke-ReID~\cite{dukemtmc-reid} and \duke-VideoReID~\cite{Wu2018ExploitTU}---are still available for download. Both \duke-ReID and \duke-VideoReID contain a cropped and edited subset of the videos from \duke{}.

Similarly, six derivatives of \msceleb{} contained subsets of or the entire original dataset. Four of these---MS1MV2~\cite{Deng2019ArcFaceAA}, MS1M-RetinaFace~\cite{Deng2019LightweightFR}, MS1M-IBUG~\cite{Deng2017MarginalLF}, and \msceleb-v1c~\cite{trillionpairs}---are still available for download. Racial Faces in the Wild~\cite{rfw} also appears available, but requires sending an email to obtain access. Further, we found that the original \msceleb{} dataset, while taken down by Microsoft, continues to be available through third-party sites such as Academic Torrents~\cite{Cohen2014}. We also identified 20 GitHub repositories that continue to make available models pre-trained on \msceleb{} data.

Clearly, one of the goals of retraction is to limit the availability of datasets. Achieving this goal requires addressing all locations where the data might already be or might become available.

\paragraph{Continued use.} Besides being available, both \msceleb{} and \duke{} have been used in numerous research papers after they were retracted in April 2019. In our sample of papers, we found that \duke{} and its derivatives had been used 73 times and \msceleb{} and its derivatives had been used 54 times in 2020. Because our samples are 20\% of our entire corpus, this equates to hundreds of uses in total. (See Figure~\ref{fig:use-over-time} for a comparison of use to previous years.)

This use further highlights the limits of retraction. Many of the cases we identified involved derivatives that were not retracted. Indeed, 72 of 73 \duke{} uses were through derivative datasets, 63 of which came from the \duke-ReID dataset, a derivative that continued to be available. Similarly, only 11 of 54 \msceleb{} uses were through the original dataset, while 17 were through derivative datasets and 26 were through pre-trained models.

One limitation of our analysis is that the use of a dataset in a paper published in 2020 (six months or more after retraction) could mean several things. The research could have been initiated after retraction, with the researchers ignoring the retraction and obtaining the data through a copy or a derivative. The research could have  begun before the retraction and the researchers may not have learned of the retraction. Or, the research could already have been under review. Regardless, it is clear that 18 months after the retractions, they have not had the effect that one might have hoped for.

\paragraph{Retractions lacked specificity and clarity.} In light of the continued availability and use of both these datasets, it is worth considering whether the retractions included sufficient information about why other researchers should refrain from using the dataset.

After the retraction, the authors of the DukeMTMC dataset issued an apology in \textit{The Chronicle}, Duke's student newspaper, noting that the data collection had violated IRB guidelines in two respects: ``Recording outdoors rather than indoors, and making the data available without protections.'' \cite{tomasi_2019} However, this explanation did not appear on the website that hosted the dataset, which was simply taken down, meaning that not all users looking for the dataset would encounter this information. The retraction of \msceleb{} fared worse: Microsoft never stated ethical motivations for removing the dataset, though the removal followed soon after multiple reports critiquing the dataset for privacy violations~\cite{megapixels}. Rather, according to reporting by \textit{The Financial Times}, Microsoft stated that the dataset was taken down ``because the research challenge is over'' \cite{murgia_2019}. The website that hosted \msceleb{} is also no longer available. Neither retraction included calls to not use the data.

The disappearance of the websites also means that license information is no longer available through these sites. We were able to locate the \duke{} license through GitHub repositories of derivatives. We were unable to locate the \msceleb{} license---which prohibits the redistribution of the dataset or derivatives---except through an archived version.\footnote{An archived version from April 2017 (found via \cite{megapixels}) is available at \url{http://web.archive.org/web/20170430224804/http://msceleb.blob.core.windows.net/ms-celeb-v1-split/MSR_LA_Data_MSCeleb_IRC.pdf}.} We discuss shortcomings of dataset licenses in Section~\ref{issue:tos}. 

\begin{table}
  \caption{Reddit posts inquiring about how to access DukeMTMC and MS-Celeb-1M after their retractions.}
  \label{table:posts-runaway}
  \vspace*{3mm}
  \footnotesize
  \centering
  \begin{tabular}{p{4cm}P{2cm}p{7cm}}
    \toprule
    % \multicolumn{2}{c}{Part}                   \\
    % \cmidrule(r){1-2}
        URL & Dataset & Post contents\\
        \midrule
        \url{https://www.reddit.com/r/computervision/comments/drg802/does_anyone_have_dukemtmc_dataset/}
        &
        DukeMTMC
        &
        Hi, currently I am working on my thesis in broad terms "Person re-Identification". All papers that tackle this problem in a way or another this dataset. It was listed as unavailable in the summer of 2019, so I am stuck in the process of finding quality data. Is there any chance that you have or know someone that has this dataset somewhere? Thanks in advance.
        \\ \\
        \url{https://www.reddit.com/r/datasets/comments/dj6zrh/duke_mtmc_alternative/}
        &
        DukeMTMC
        &
        As of June 2019, the Duke MTMC surveillance dataset was discontinued following a privacy investigation by the financial times. Does anyone know of an alternative source to download it from, or just an alternative dataset all together?
        \\ \\
        \url{https://www.reddit.com/r/datasets/comments/fpvs47/does_anyone_have_approach_to_get_dukemtmc_dataset/}
        &
        DukeMTMC
        &
        Does anyone have approach to get DukeMTMC dataset? if not, please recommend some other pedestrian datasets in MTMC. Thanks a lot!
        \\ \\
        \url{https://www.reddit.com/r/datasets/comments/cvq6sa/download_raw_msceleb1m/}
        &
        MS-Celeb-1M
        &
        Hi, I need to download the original MS-Celeb-1M (academic purposes). I tried on megapixels but I could not find any link. There is a reference to a clean dataset (https://github.com/PINTOFSTU/C-MS-Celeb), which, in fact, its only a label list. At academic torrents there is a torrent file, but this dataset is not the original, the images are already croped or aligned. Thanks in advance,
        \\ \\
        \url{https://www.reddit.com/r/DataHoarder/comments/bxz19f/anyone_have_it_microsoft_takes_down_huge/}
        &
        MS-Celeb-1M
        &
        Anyone have it?: Microsoft takes down huge MS-Celeb-1M facial recognition database
        \\
    \bottomrule
  \end{tabular}
\end{table}

We also identified public efforts to access and preserve these datasets, perhaps indicating confusion about the substantive meaning of the dataset's retractions. We found three and two Reddit posts inquiring about the availability of \duke{} and \msceleb{}, respectively, following their retraction. Two of these posts (one for each dataset) noted or referred to investigations about potential privacy violations, but still inquired about where the dataset could be found. These posts are listed in \Cref{table:posts-runaway}.

In contrast to the retractions of \duke{} and \msceleb{}, the retraction of TinyImages was more clear. On the dataset's website, the creators ask that ``the community to refrain from using it in future and also delete any existing copies of the dataset that may have been downloaded''~\cite{tinyimageswebsite}.

\section{Supplement: Analysis of license restrictions}

We describe findings summarized in \Cref{table:tos} about the effectiveness of license restrictions for mitigating harms.

\paragraph{Licenses do not effectively restrict production use.}

We analyzed the licensing information for DukeMTMC, MS-Celeb-1M, and LFW, and determined the implications for production use. Datasets are at a greater risk to do harm in production settings, where characteristics of a dataset directly affect people. 

DukeMTMC is released under the CC BY-NC-SA 4.0 license, meaning that users may freely share and adapt the dataset, as long as attribution is given, it is not used for commercial purposes, derivatives are shared under the same license, and no additional restrictions are added to the license. Benjamin et al. \cite{montreal} note many possible ambiguities in a ``non-commercial'' designation for a dataset. We emphasize, in particular, that this designation allows the possibility for non-commercial production use. Models deployed by nonprofits and governments maintain risks associated with commercial models. Additionally, there is  legal ambiguity regarding whether models trained on the data may be used for commercial purposes. 

MS-Celeb-1M is released under a Microsoft Research license agreement,\footnote{The license is no longer publicly available. An archived version is available here: \url{http://web.archive.org/web/20170430224804/http://msceleb.blob.core.windows.net/ms-celeb-v1-split/MSR_LA_Data_MSCeleb_IRC.pdf}} which has several stipulations, including that users may ``use and modify this Corpus for the limited purpose of conducting non-commercial research.'' Again, implications for commercial use of pre-trained models may be ambiguous.

LFW was  released without any license. In 2019, a disclaimer was added on the dataset's website, indicating that the dataset ``should not be used to conclude that an algorithm is suitable for any commercial purpose'' \cite{lfw-web}. The lack of an original license meant that the dataset's use was entirely unrestricted until 2019. Furthermore, while it includes useful guiding information, the disclaimer does not hold legal weight. Additionally, through an analysis of results given on the LFW website \cite{lfw-web}, we found four commercial systems that clearly advertised their performance on the datasets,
% \footnote{\textit{Innovative Technology} (\url{https://www.innovative-technology.com/icu}) — ``Using our own AI algorithms developed over many years, ICU offers an accurate (99.88\%*) precise and affordable facial recognition system *Source: LFW''; \textit{Oz Forensics} (\url{https://ozforensics.com/#main_window}) — ``The artificial intelligence algorithms recognize people with 99.87\% accuracy''; \textit{IntelliVision} (\url{https://www.intelli-vision.com/facial-recognition/}) — ``Facial recognition accuracy over 99.5\% on public standard data sets. It scores the following accuracy in the leading public test databases – LFW: 99.6\%, YouTube Faces: 96.5\%, MegaFace (with 1000 people/distracters): 95.6\%''; \textit{CyberExtruder} (\url{https://cyberextruder.com/aureus-insight/}): ``To that end, test results from well-known, publicly-available, industry standard data sets including NIST’s FERET and FRGC tests and the UMass Labeled Faces in the Wild data set are shown below.''} 
though we do not know if the disclaimer is intended to discourage this behavior:
\begin{itemize}[leftmargin=*]
    \item \textbf{Innovative Technology.} \url{https://www.innovative-technology.com/icu} \\ ``Using our own AI algorithms developed over many years, ICU offers an accurate (99.88\%*) precise and affordable facial recognition system *Source: LFW''
    \item \textbf{Oz Forensics.} \url{https://ozforensics.com/#main_window} \\ ``The artificial intelligence algorithms recognize people with 99.87\% accuracy.''
    \item \textbf{IntelliVision.} \url{https://www.intelli-vision.com/facial-recognition/} \\ ``Facial recognition accuracy over 99.5\% on public standard data sets. It scores the following accuracy in the leading public test databases – LFW: 99.6\%, YouTube Faces: 96.5\%, MegaFace (with 1000 people/distracters): 95.6\%.''
    \item \textbf{CyberExtruder.} \url{https://cyberextruder.com/aureus-insight/} \\ ``To that end, test results from well-known, publicly-available, industry standard data sets including NIST’s FERET and FRGC tests and the UMass Labeled Faces in the Wild data set are shown below.''
\end{itemize}
Because LFW is a relatively small dataset, its use as training data in production settings is unlikely. Risk remains, however, as the use of its performance as a benchmark on commercial systems can lead to overconfidence both among the system creators and potential clients.

ImageNet's ``terms of access'' specifies that the user may use the database ``only for non-commercial research and educational purposes.'' Again, implications for commercial use of pre-trained models may be ambiguous.

\paragraph{Derivatives do not always inherit original terms.}
DukeMTMC, MS-Celeb-1M, and ImageNet---according to their licenses---may only be used for non-commercial purposes. We analyzed available derivatives of each dataset to see if they include a non-commercial use designation. All four DukeMTMC derivative datasets included the designation. Four of seven MS-Celeb-1M derivative datasets included the designation. Only three of 21 repositories containing models pre-trained on MS-Celeb-1M included the designation. We also identified 12 repositories containing models pre-trained on ImageNet, of which only three restricted commercial use. Furthermore, Keras, PyTorch, and MXNet all come built in with numerous models pre-trained on ImageNet, and are licensed for commercial use. (This analysis does not apply to LFW, which was released with no license.)

Thus, we found mixed results of license inheritance. We note that DukeMTMC's license specifies that derivatives must include the original license. Meanwhile, MS-Celeb-1M's license, which prohibits derivative distribution in the first place, is no longer publicly accessible, perhaps partially explaining our findings. Licenses are only effective if actively followed and inherited by derivatives.

The loose licenses associated with the pre-trained models are particularly notable. Of the 21 repositories containing models pre-trained on MS-Celeb-1M, seven contained the MIT license, one contained the Apache 2.0 license, and one contained the BSD-2-Clause. Each of these licenses permit commercial use. Additionally, nine repositories were released with no license at all.

\section{Supplement: Posts discussing legality of pre-trained models}
As discussed in \Cref{issue:tos}, we identified 14 posts discussing the legality of using models pre-trained on a non-commercial dataset for commercial purposes. We list these posts in \Cref{table:license-discussions}. We identified these posts via four Google searches with the query ``pre-trained model commercial use.'' We then searched the same query on Google with ``site:www.reddit.com,'' ``site:www.github.com,''  ``site:www.twitter.com,'' and ``site:www.stackoverflow.com.'' These are four sites where questions about machine learning are posted. For each search, we examined the top 10 sites presented by Google. Within relevant posts, we also extracted any additional relevant links included in the discussion.

\begin{table}
  \caption{Discussion posts about the legality of the commercial use of models pre-trained on non-commercial data.}
  \label{table:license-discussions}
  \vspace*{3mm}
  \centering
  \scriptsize
  \begin{tabular}{p{1.6cm}>{\raggedright}p{1.9cm}p{9.2cm}}
    \toprule
        \textbf{Discussion site} & \textbf{Dataset discussed}  &  \textbf{URL}\\
        \midrule
        GitHub &	ImageNet &	\url{https://github.com/keras-team/keras-applications/issues/140}\\
GitHub	&ImageNet	&\url{https://github.com/keras-team/keras/issues/13362}\\
GitHub	&ImageNet	&\url{https://github.com/pytorch/vision/issues/2597}\\
GitHub	&ImageNet	&\url{https://github.com/tensorflow/models/issues/9131}\\
GitHub	&LIP	&\url{https://github.com/Engineering-Course/LIP_JPPNet/issues/42}\\
Twitter	&ImageNet	&\url{https://twitter.com/viglovikov/status/1296292326478761984}\\
Kaggle	&ImageNet	&\url{https://www.kaggle.com/c/deepfake-detection-challenge/discussion/131121}\\
Reddit	&ImageNet	&\url{https://www.reddit.com/r/deeplearning/comments/9lalpv/using_pretrained_deep_neural_networks_for/}\\
Reddit	&ImageNet	&\url{https://www.reddit.com/r/MachineLearning/comments/4eu2vd/can_pretrained_networks_be_used_in_commercial/}\\
Reddit	&ImageNet	&\url{https://www.reddit.com/r/MachineLearning/comments/7eor11/d_do_the_weights_trained_from_a_dataset_also_come/}\\
Reddit	&ImageNet	&\url{https://www.reddit.com/r/MachineLearning/comments/id4394/d_is_it_legal_to_use_models_pretrained_on/}\\
Reddit	&Scene Flow	&\url{https://www.reddit.com/r/MLQuestions/comments/hwxfth/if_i_have_a_dataset_whose_license_restricts_me/}\\
Reddit	&FFHQ	&\url{https://www.reddit.com/r/MachineLearning/comments/lzmm75/d_legal_mess_in_ml_datasetspretrained_modelscode/}\\
MXNet	&ImageNet	&\url{https://discuss.mxnet.apache.org/t/commercial-use-license-for-pre-trained-models/3343}\\
    \bottomrule
  \end{tabular}
\end{table}

\section{Supplement: Dataset management and citation}\label{app:citation}

In \Cref{issue:citation}, we showed how dataset management and citation can help mitigate harms through facilitating documentation, transparency and accountability, and tracking, and summarized findings showing how current practices fall short in achieving these aims. We present these findings in detail below.

\paragraph{Dataset management practices raise concerns for persistence.} Whereas other academic fields utilize shared repositories,\footnote{\emph{Nature} provides specific guidance for both field-specific and general data repositories (\url{https://www.nature.com/sdata/policies/repositories}).} machine learning datasets are often managed through the websites of individual researchers or academic groups. None of the 38 datasets in our analysis are managed through shared repositories. Unsurprisingly, we found that some datasets were no longer maintained (which is different from being retracted).

We were only able to find information about D31 and D38 through archived versions of sites found via the Wayback Machine. And even after examining archived sites, we were unable to locate information about D6, D17, and D25. Another consequence is the lack of persistence of documentation. Ideally, information about a dataset should remain available even if the dataset itself is no longer available. But we found that after DukeMTMC and MS-Celeb-1M were taken down, so too were the sites that contained their terms of use.

\begin{table}[t]
\caption{Examples of dataset references that were challenging to disambiguate.}
\label{table:citation-disambiguation}
\vspace*{3mm}
\centering
\footnotesize
\begin{tabular}{p{6.5cm}p{6.5cm}}\toprule

\multicolumn{1}{c}{\textbf{Reference}} & \multicolumn{1}{c}{\textbf{Attempted disambiguation}} \\
\midrule
\scriptsize
``Experiments were performed on four of the largest ReID benchmarks, i.e., Market1501 [45], CUHK03 [17], DukeMTMC [33], and MSMT17 [40] … DukeMTMC provides 16,522 bounding boxes of 702 identities for training and 17,661 for testing.'' & \scriptsize Here, the dataset is called DukeMTMC and the citation [33] is of DukeMTMC’s associated paper. However, the dataset is described as an ReID benchmark. Moreover, the statistics given exactly match the popular DukeMTMC-ReID derivative (an ReID benchmark). This leads us to believe DukeMTMC-ReID was used.\\
\\
\scriptsize``We used the publicly available database Labeled Faces in the Wild (LFW)[6] for the task. The LFW database provides aligned face images with ground truth including age, gender, and ethnicity labels.'' & \scriptsize The name and reference both point to the original LFW dataset. However, the dataset is described to contain aligned images with age, gender, and ethnicity labels. The original dataset contains neither aligned images nor any of these annotations. There are, however, many derivatives with aligned versions or annotations by age, gender, and ethnicity. Since no other description was given, we were unable to disambiguate.\\
\\
\scriptsize``MS-Celeb-1M includes 1M images of 100K
subjects. Since it contains many labeling noise, we use a cleaned version of MS-Celeb-1M [16].'' & \scriptsize The paper uses a “cleaned version of MS-Celeb-1M,” but the particular one is not specified. (There are many cleaned versions of the dataset.) The citation [16] is to the original MS-Celeb-1M’s associated paper and no further description is given. Therefore, we were unable to disambiguate.\\
\bottomrule
\end{tabular}
\end{table}

\paragraph{Dataset references can be difficult to disambiguate.} Clear dataset citation is important for harm mitigation. However, datasets are not typically designated as independent citable research objects like academic papers are. This is evidenced by a lack of standardized permanent identifiers, such as DOIs. None of the 38 datasets we encountered in our analysis had such identifiers. Datasets are often assigned DOIs when added to shared data repositories.

Without dataset-specific identifiers, we found that datasets were typically cited with a combination of the dataset’s name, a description, and paper citations. In many cases, an associated paper is cited---a paper through which a dataset was introduced or that the dataset’s creators request be cited. In some cases, a dataset does not have a clear associated paper. For example, D31 was not introduced in an academic paper and D20’s creators suggest three distinct academic papers that may be cited. This practice can lead to challenges in identifying and accessing the dataset(s) used in a paper, especially when the name, description, or citation conflict. There is a discrepancy between the roles of citation for attribution and documentation: providing sufficient attribution does not necessarily imply that sufficient documentation is given, and vice versa.

In our analysis, 42 papers included dataset references that we were unable to fully disambiguate. Oftentimes, this was a result of conflating a dataset with its derivatives. For example, we found nine papers that suggested that images in LFW were annotated with attributes or keypoints, but did not specify where these annotations were obtained. (LFW only contains images labeled with identities and many derivatives of LFW include annotations.) Similarly, seven papers indicated that they used a cleaned version of MS-Celeb-1M, but did not identify the particular derivative. We were able to disambiguate the references in 404 papers using a dataset or a derivative, but in many of these instances, making a determination was not direct (for instance, see the first example in Table~\ref{table:citation-disambiguation}).

\paragraph{Datasets and documentation are not directly accessible from citations.} We found that accessing datasets from papers is not currently straightforward. While data access requirements, such as sections dedicated to specifying where datasets and other supporting materials may be found, are common in other fields, they are rare in machine learning. We sampled 60 papers from our sample that used DukeMTMC, MS-Celeb-1M, LFW, or one of their derivative datasets, and only six provided access information (each as a URL).

Furthermore, the descriptors we mentioned above---paper citations, name, and description---do not offer a direct path to the dataset. The name of a dataset can sometimes be used to locate the dataset via web search, but this works poorly in many instances---for example, when a dataset is not always associated with a particular name or when the dataset is not even available. Datasets D27, D28, D31, D32, and D38 are not named. In other cases, datasets may be known by multiple names. Equating datasets can be challenging. As one GitHub user commented: ``Since there are many different names regarding different versions of ms1m dataset, below is my own understanding for these different names:
ms1m-v1 = ms1m-ibug[,]
ms1m-v2 = ms1m-arcface[,]
both of them are detected by mtcnn and use the same alignment procedure. Am I understanding correctly?''\footnote{\url{https://github.com/deepinsight/insightface/issues/513} Another post expressing similar confusion about MS-Celeb-1M derivatives is available here: \url{https://github.com/deepinsight/insightface/issues/566}.}. Here, note that ``ms1m'' is a common abbreviation for MS-Celeb-1M.

Citations of an associated paper also do not directly convey access information. As an alternate strategy, we were able to locate some datasets by searching for personal websites of the dataset creators or of their associated academic groups. However, as we mentioned earlier, we were still unable to locate D6, D17, and D25, even after looking at archived versions of sites.

\begin{figure}
    \centering
    \includegraphics[width=0.8\linewidth]{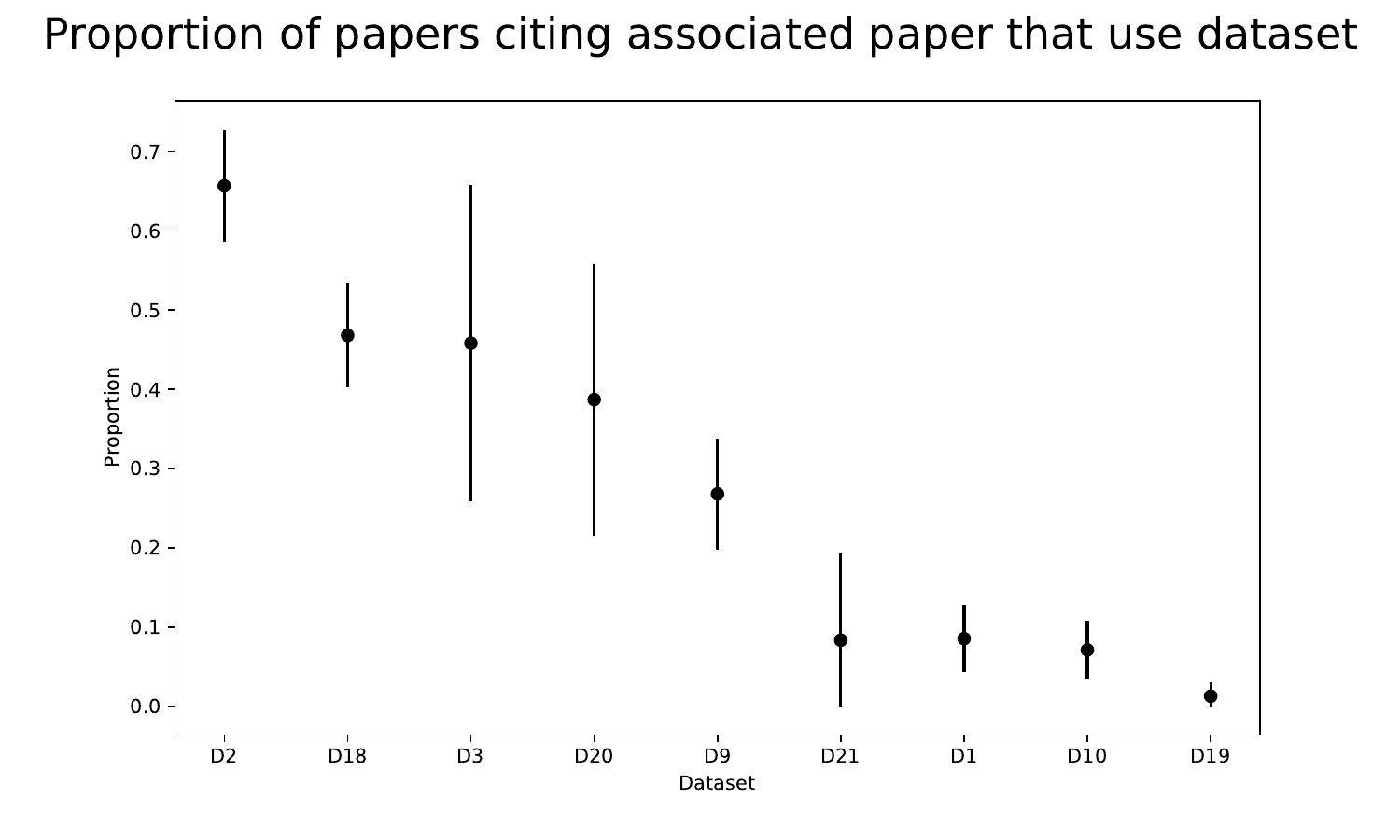}
    \caption{Papers citing associated papers often do not use the associated dataset. The proportion that do varies greatly across different datasets. Here, we include associated papers for which we sampled at least 20 citing papers, and show 95 percent confidence intervals.}
    \label{fig:citation-proportion}
\end{figure}

\paragraph{Current infrastructure makes tracking dataset use difficult.} A lack of dataset citations also makes it difficult to track dataset use. Citations of associated papers are not necessarily effective proxies in this respect. On one hand, the proportion of papers citing an associated paper that use the corresponding dataset varies significantly (see Figure~\ref{fig:citation-proportion}). This is because papers citing an associated paper may be referencing other ideas mentioned by the paper. On the other hand, some datasets may be commonly used in papers that do not cite a particular associated paper. Of the papers we found to use DukeMTMC-ReID, 29 cited the original dataset, 63 cited the derivative dataset, and 50 cited both. Furthermore, some datasets may not have a clear associated paper, and various implementations of pre-trained models are unlikely to have associated papers. Thus, associated papers---as currently used---are an exceedingly clumsy way to track the use of datasets.

Tracking derivative creation presents an even greater challenge. Currently, there is no clear way to identify all derivatives of a dataset. The websites of LFW and DukeMTMC (the latter no longer online), maintained lists of derivatives. However, our analysis reveals that these lists are far from complete. Proponents of dataset citation have suggested the inclusion of metadata indicating provenance in a structured way (thus linking a dataset to its derivatives) \cite{Groth2020FAIRDR}, but such a measure has not been adopted by the machine learning community.

Ambiguities in dataset citation and the instability of datasets present fundamental challenges to alternative approaches to automating the tracking of dataset use and derivative creation. Meanwhile, the adoption of standard practices in dataset management and citation can enable both of these tasks.

\section{Supplement: Identifying models pre-trained on MS-Celeb-1M and ImageNet}

In our analysis, we identified GitHub repositories containing models pre-trained on MS-Celeb-1M and ImageNet. We describe our methodology below.

For each of the six pre-trained ``model classes’’ we identified in our analysis (all pre-trained on MS-Celeb-1M), we identified if and where the corresponding pre-trained models are available on GitHub. We first identified repositories linked in the papers using the model. Within these repositories, we also examined any linked third-party implementations. We further searched the name of the model class on GitHub and examined the first 10 results for if they contained a model pre-trained on MS-Celeb-1M. This resulted in a total of 21 repositories (listed in \Cref{table:msceleb1m-pretrained}), 20 of which currently contain a model pre-trained on MS-Celeb-1M.

\begin{table}
  \caption{GitHub repositories of models pre-trained on MS-Celeb-1M}
  \label{table:msceleb1m-pretrained}
  \vspace*{3mm}
  \centering
  \scriptsize
  \begin{tabular}{p{0.3cm}p{0.3cm}p{3cm}p{8cm}}
    \toprule
        \rot{\textbf{Model class}} & \rot{\textbf{Still available}}  & \rot{\textbf{License}}  &  \rot{\textbf{GitHub url}}\\
        \midrule
M1	&\checkmark	&MIT	&\url{https://github.com/cydonia999/VGGFace2-pytorch} \\
M1	&	&none	&\url{https://github.com/ox-vgg/vgg_face2} \\
M2	&\checkmark	&MIT	&\url{https://github.com/seasonSH/Probabilistic-Face-Embeddings} \\
M3	&\checkmark	&non-commercial research	&\url{https://github.com/deepinsight/insightface/tree/master/recognition/arcface_torch} \\
M3	&\checkmark	&MIT	&\url{https://github.com/auroua/InsightFace_TF} \\
M3	&\checkmark	&MIT	&\url{https://github.com/AIInAi/tf-insightface} \\
M3	&\checkmark	&MIT	&\url{https://github.com/TreB1eN/InsightFace_Pytorchh} \\
M3	&\checkmark	&none	&\url{https://github.com/ronghuaiyang/arcface-pytorch} \\
M3	&\checkmark	&none	&\url{https://github.com/gehaocool/CombinedMargin-caffe} \\
M3	&\checkmark	&none	&\url{https://github.com/luckycallor/InsightFace-tensorflow} \\
M3	&\checkmark	&MIT	&\url{https://github.com/wang-xinyu/tensorrtx} \\
M3	&\checkmark	&none	&\url{https://github.com/deepinsight/insightface/wiki/Model-Zoo} \\
M3	&\checkmark	&Apache 2.0	&\url{https://github.com/foamliu/InsightFace-PyTorch} \\
M3	&\checkmark	&non-commercial research	&\url{https://github.com/leondgarse/Keras_insightface} \\
M4	&\checkmark	&none	&\url{https://github.com/AlfredXiangWu/LightCNN} \\
M4	&\checkmark	&non-commercial	&\url{https://github.com/AlfredXiangWu/face_verification_experiment} \\
M4	&\checkmark	&none	&\url{https://github.com/yxu0611/Tensorflow-implementation-of-LCNN} \\
M4	&\checkmark	&none	&\url{https://github.com/lyatdawn/LightCNN-mxnet} \\
M5	&\checkmark	&MIT	&\url{https://github.com/davidsandberg/facenet/tree/accd6881d58b3bf7bfbdc12bae2d6dde738ba48e} \\
M5	&\checkmark	&none	&\url{https://github.com/nyoki-mtl/keras-facenet} \\
M6	&\checkmark	&BSD-2-Clause	&\url{https://github.com/penincillin/DREAM} \\
    \bottomrule
  \end{tabular}
\end{table}

We performed a similar search to identify models pre-trained on ImageNet. For this, we searched ``ImageNet pretrained'' on GitHub and then examined the first 20 results. This yielded 12 repositories that contained a model pre-trained on ImageNet, listed in \Cref{table:imagenet-pretrained}.

\begin{table}
  \caption{GitHub repositories of models pre-trained on ImageNet}
  \label{table:imagenet-pretrained}
  \vspace*{3mm}
  \centering
  \scriptsize
  \begin{tabular}{>{\raggedright}p{3cm}p{8cm}}
    \toprule
        \textbf{License}  &  \textbf{GitHub url}\\
        \midrule
none	&\url{https://github.com/PengBoXiangShang/MobileNetV3_PyTorch}\\
none	&\url{https://github.com/qubvel/resnet_152}\\
none	&\url{https://github.com/visionNoob/pytorch-darknet19}\\
MIT	&\url{https://github.com/flyyufelix/DenseNet-Keras}\\
MIT	&\url{https://github.com/qubvel/segmentation_models.pytorch}\\
MIT	&\url{https://github.com/Alibaba-MIIL/ImageNet21K}\\
non-commercial research	&\url{https://github.com/HiKapok/TF-SENet}\\
non-commercial research	&\url{https://github.com/HiKapok/Xception_Tensorflow}\\
BSD-3-Clause	&\url{https://github.com/Cadene/pretrained-models.pytorch}\\
Apache-2.0	&\url{https://github.com/kuan-wang/pytorch-mobilenet-v3}\\
Apache-2.0	&\url{https://github.com/pudae/tensorflow-densenet}\\
Creative Commons Attribution-NonCommercial-ShareAlike 4.0 International Public License for Noncommercial use only	&\url{https://github.com/Res2Net/Res2Net-PretrainedModels}\\
    \bottomrule
  \end{tabular}
\end{table}

\section{Recommendations: The role of the IRB}\label{app:recommendations}

In \Cref{sec:recommendations}, we outlined recommendations for several stakeholders. In particular, we suggested that PCs take a larger role in regulating dataset use and creation. Here, we address the role of Institutional Review Boards (IRBs), which have historically played a fundamental role in regulating research ethics.

Researchers have recently called for greater IRB oversight in dataset creation \cite{pyrrhic}, and IRBs have certain natural advantages in regulating datasets. IRBs may have more ethics expertise than program committees; IRBs are also able to review datasets prior to their creation. Thus, IRBs can prevent harms that occur during the creation process. 

However, conceived first to address biomedical research, IRBs have been an imperfect fit for data-centered research. Notably ``human subjects research” has a narrow definition and thus many of the datasets (and associated research) that have caused ethical concern in machine learning would not fall under the purview of IRBs. An even more significant limitation is that IRBs are not allowed to consider downstream harms \cite{metcalf_2017}.\footnote{``The IRB should not consider possible long-range effects of applying knowledge gained in the research (e.g., the possible effects of the research on public policy) as among those research risks that fall within the purview of its responsibility'' (45 CFR §46.111).}

Unless and until the situation changes, our primary recommendation regarding IRBs is for researchers to recognize that research being approved by the IRB does not mean that it is ``ethical,'' and for IRBs themselves to make this as clear as possible.

\end{document}